# Bit Efficient Quantization for Deep Neural Networks

Prateeth Nayak  David Zhang  Sek Chai
Latent AI  SRI International  Latent AI

*Abstract*— Quantization for deep neural networks have afforded models for edge devices that use less on-board memory and enable efficient low-power inference. In this paper, we present a comparison of model-parameter driven quantization approaches that can achieve as low as 3-bit precision without affecting accuracy. The post-training quantization approaches are data-free, and the resulting weight values are closely tied to the dataset distribution on which the model has converged to optimality. We show quantization results for a number of state-of-art deep neural networks (DNN) using large dataset like ImageNet. To better analyze quantization results, we describe the overall range and local sparsity of values afforded through various quantization schemes. We show the methods to lower bit-precision beyond quantization limits with object class clustering.

## I. INTRODUCTION

Significant progress has been recently made on algorithms, networks and models to enable low-power edge applications. For example, advances in network architecture search [1], parameter quantization [2], and pruning [3] have afforded models to below 100kB. However, there is still active research to understand the relationships among low bit-precision, data representation, and neural model architecture.

It is well understood in the deep learning community that to capture a wide spectrum of low, mid and high-level features for deep semantic understanding of complex patterns, DNNs with many layers, nodes, and with high local and global connectivity are needed. The success of recent DNNs (e.g. ResNet, Yolo, MobileNet) in speech, vision, and natural language, comes in part from the ability to train much larger models on much larger dataset [4-6]. One fundamental challenge for quantization is to find best assignment of bits for DNN parameter values that have a degree of non-linearity that increases exponentially as dataset size increases.

Through quantization, parameters for DNN can be converted from 32-bit floating point (FP32) towards 16-bit or 8-bit models, with minimal loss of accuracy. There are even techniques that can quantized down to binary (single bit precision). The main benefit of moving to lower bit-precision is the higher power efficiency and smaller storage needed. Moreover, DNN inference can be processed using more efficient and parallel processor hardware. And as such, quantization makes it possible to run DNN workloads that support object and voice recognition, navigation, and medical image analysis on smartphones and other edge devices.

Through training, the DNN model is learning a representation that support tasks such as classification. The converged DNN parameter values (e.g. its range and distribution) represent the relationship between filter types and class distribution. Therefore, in order to maintain algorithmic performance after quantization, it is desirable to maintain a "similar" range and distribution of learned DNN representations in the quantized parameter values.

In this paper, we present a comparison of data-free quantization schemes (i.e. asymmetric, symmetric, logarithmic) on a number of DNN models trained with large dataset such as ImageNet. Similar to [8], our quantization approaches do not need training data for calibration. Instead, we show model-parameter driven approaches can achieve as low as 3-bit precision without significantly affecting accuracy. Furthermore, we explore the quantization approach based on the overall range and local sparsity of values to provide insights to how well we can shrink a DNN model for a given training dataset. We describe how well bits represent the model after parameter quantization in preserving local distribution and a broader range-per-bit for compression.

Our goal is to directly address the understanding of quantization limits by exploring inter-layer comparison of range value and distribution. In fact, we find that for a large dataset like ImageNet, average bit-precision is 6-bits per parameter, for a number of state-of-art DNN, adding to previously published results [8-10]. Using range and distribution of the quantized parameter values, we can better describe quantization effects and offer insights for further compression.

The research area in efficient DNN inference, particularly through quantization, is gaining momentum and moving at a very fast paced. To the best of our knowledge, we offer the following contributions in this paper:

- Data-free quantization approaches, including (1) uniform asymmetric, (2) uniform symmetric, and (2) logarithmic using power-of-two, with results below 8-bit precision
- Analysis of parameter range and distribution across DNN layers, with results of DNN models for ImageNet
- An approach using object class clustering to lower bit-precision beyond quantization limits
- Comparison of accuracy and memory savings using better file-compression (Gzip and 7.zip), with analysis of encoding scheme.

This paper is organized as follows: in Section 2, we present comparisons against related research in DNN quantization. In Section 3, we briefly define our quantization approaches with highlights on the range and distribution. In Section 4, we describe our simulation setup, and we provide early results and associated analysis that evaluates our approach. Finally, in Section 5, we present our conclusions and discuss our future work in this space.

## II. Related Research

In recent DNN quantization methods, FP32 values are compressed onto a regularly spaced grid, with the original values approximated with fixed point representations (e.g. integer values) [7]. However, many implementations of this quantization approach require the selection of a scaling factor (e.g. max and min ranges) and zero-point offset to allow such a mapping [2]. Other related quantization approaches have been published recently [11-13] for reducing bit-precision and memory footprint.

Our quantization approaches do not need training data for calibration. Based on the quantization taxonomy provided in [8], our approach can be considered a Level-1 post-training (No data and no backpropagation required. Method works for any model). Similar to [8], our approach is a data-free quantization, and in this paper, we show model-parameter driven approaches that can achieve as low as 3-bit precision without significantly affecting accuracy. Furthermore, we explore the quantization's approach based on the overall range and local sparsity of values to provide insights to how well we can shrink a DNN model for a given training dataset.

Other methods for quantization need architecture changes or training with quantization in mind [9-10,14-16]. These methods may include network architecture search to selectively bound the target bit precision or fine-tuning the bit-precision iteratively during training. Similarly, training aware quantization methods may binarize [17-19] or ternarize [10] the resultant networks so that they operate at great efficiency for inference (e.g. expensive multiplications and additions are replaced by logical operations). Training aware quantization incur a larger overhead to arrive at quantized parameters because they generally need to be trained from scratch. Furthermore, quantizing models to binary often leads to strong performance degradation.

Even though these approaches share the same objective, the methods are quite different. In this paper, we focus on post-training quantization using data-free quantization schemes (i.e. uniform asymmetric, logarithmic). Many previous papers published results at byte precision. To better offer qualitative comparisons of quantization approaches, we show results below the 8-bit boundaries to find insights on bit-precision limits. Furthermore, previous quantitative comparisons of these approaches are focused on overall file-size vs accuracy, rather than the fitting to the overall inter-level distributions of quantized parameter values from the original FP32 distribution.

## III. Approach

In this section, we provide the details on the different quantization approaches which are purely model parameter driven. First, we provide the mathematical formulation for these approaches. Then we describe the quantization effects by observing that ability of the approach to preserve the range and density of parameter values. Then we show visualization of the quantization effects on the model parameter values from the full-precision to lower precision of a selected DNN layer.

### III.a. Quantization Schemes

We have selected two linear, range-based approach (asymmetric, and symmetric) and one logarithmic-scale (power-of-two) approach. The approaches are chosen to give different benefits on the same tensor quantization.

(1) **Uniform-ASYMM** - In this mode, we map the minimum and maximum of the float range to integer range with a quantization bias term. Let us denote the original tensor by $x_f$, output quantized tensor by $x_q$, and chosen number of bits for quantization by $n$.

$$x_q = round\left(\left(x_f - min_{x_f}\right)\frac{2^n - 1}{max_{x_f} - min_{x_f}}\right)$$

(2) **Uniform-SYMM** – In this mode, we map original tensor to the quantized range with maximum absolute value of the minimum or maximum of the original tensor. There is no quantization bias term and it is symmetric around zero.

$$x_q = round\left(x_f \frac{2^{n-1} - 1}{max|x_f|}\right)$$

(3) **Power-of-2** – This mode involves mapping the values original tensor to the closest power of 2 value. This preserves the range of float tensor.

$$x_q = (signmask)2^{round(log_2|x_f|)}$$

For evaluation, we used the workflow [21] that builds upon a TensorFlow framework to implement these quantization schemes. The workflow supports the import of various convolution-based models and we quantize the model parameters, namely, convolution kernels, batch norm layers, and dense layers. The resulting model post-quantization is evaluated within the same framework for a given set of test data. It is to be noted that, the data is only required for the purpose of evaluation and not for the quantization of the model itself.

### III.b. Quantization Effects on Weight Distribution

Most researchers consider accuracy drop [2,3,7,8] as the only metric to evaluate the performance of the quantization method. Instead, we observed that quantization results can be characterized on how well the model parameter values are spread on the quantized range. That is, given a bit-precision target, quantization approaches are effectively striving to achieve a Bit-Efficiency (BE) threshold. A quantization result is said to have a better BE, if the method can provide a higher range-per-bit coverage of baseline precision whilst preserving the density-per-bit for each of the quantized level as compared to the baseline. Much like a bit efficiency metric in compression standards (i.e. best utilization of a bit to encode a data stream), neural network quantization should have a similar BE comparison to more eloquently describe the quantization result.

Figure 1 illustrates the quartile range plots per-channel for different quantization results on the 9th layer (32 channels) of ResNet18 pre-trained on CIFAR10. Here, we can see that the uniform quantization approaches are able to preserve the mean of the baseline tensor, while the logarithmic approach of power-of-two maintains the range for each channel with lesser outlier parameters.

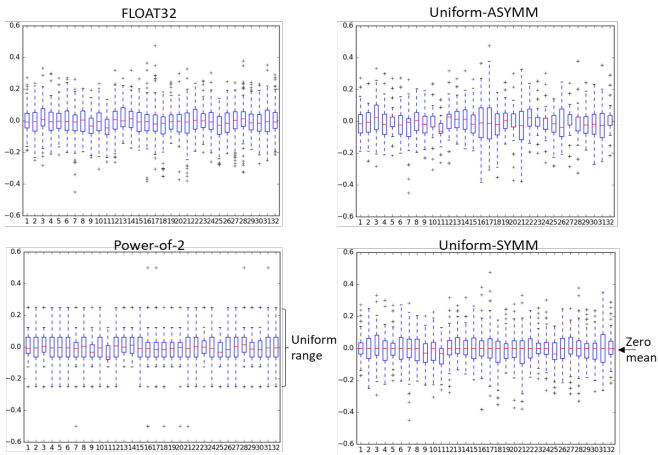

Figure 1. Quantization effects on quartile ranges of Tensors for different approaches. The FLOAT32 is the original tensor.

Figure 2 (left) shows the histogram distribution of weights from different quantization approach and Figure 2 (right) shows the weight histogram distribution from 32 to 2 bit-quantization. Regardless of the approach, the distribution profile is close. We find that quantization results are similar if the assumption that range and density of the model parameters are preserved. The *best* quantization method is the one that is able to preserve the original model parameter distribution. As such, we are not advocating a single quantization approach, but instead, a workflow that provides the flexibility to achieve the best BE. Furthermore, beyond BE, selected quantization schemes (such as power-of-two) may provide flexibility in hardware performance (discussed later in Section IV).

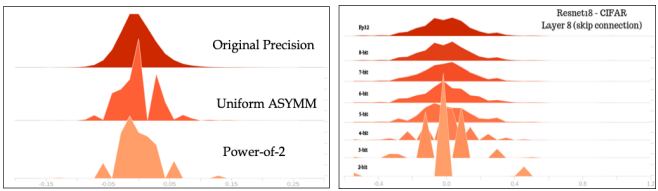

Figure 2. Histogram of weight distribution of ResNet Layers for bit precisions and approaches.

## IV. RESULTS AND EVALUATION

We evaluated the three different quantization schemes using state-of-art baseline models and dataset. First, we show results using CIFAR10 classification task to highlight opportunities to improve accuracy with object class clustering. Second, we show quantization results for six classification models on ImageNet classification task. We iteratively evaluate these models after quantizing each level of bit-precision. Third, we include quantization results for two models pre-trained on PASCAL-VOC detection task, and MS-COCO detection task. Finally, we show compression effects and power-savings from quantization using power-of-two.

IV.a. Hierarchical Clustering to Improve Accuracy

Table 1 shows quantization results for ResNet-18 pre-trained for CIFAR10 classification achieving as low as 3-bit precision without affecting accuracy drastically. We show results at byte boundary for comparison purposes with published results.

| Quantization Approach / Bit-Width | FP-32 | 8-bits | 7-bits | 6-bits | 5-bits | 4-bits | 3-bits | 2-bits |
|---|---|---|---|---|---|---|---|---|
| **Uniform-ASYMM** (Acc %; Size Mb) | 90.87 (2) | 90.83 (1.4) | 90.80 (1.35) | 90.90 (1.30) | 90.61 (1.25) | 90.04 (1.20) | 85.93 (1.16) | 22.46 (1.11) |
| **Uniform-SYMM** (Acc %; Size Mb) | 90.87 (2) | 90.51 (0.76) | 90.46 (0.65) | 90.48 (0.55) | 90.41 (0.45) | 89.54 (0.366) | 88.93 (0.291) | 77.24 (0.223) |
| **POWER-OF-2** (Acc %; bit levels) | 86.9 (Avg. Bit-Levels = 4) | | | | | | | |

Table 1. ResNet18 Quantization Results down-to 2-bits with model size.

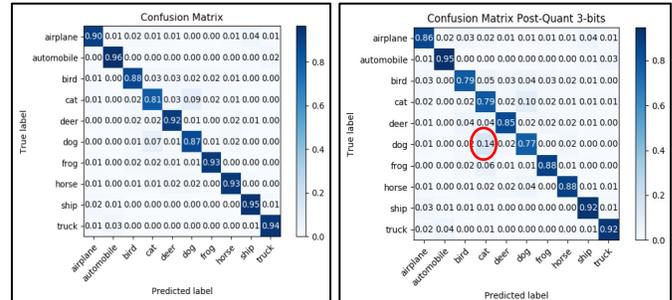

Figure 3. Confusion matrices at baseline and 3-bit. The classes *dog* and *cat* are misclassified due to overlapping distribution characteristics. This effect of quantization degrades model performance at lower bit-precision.

In this example, we observed in the confusion matrix (Figure 3) that quantization at low-bit precisions causes closely overlapping class distributions to be misclassified. This is due the effect of reducing the differentiating ability with the lower bit-precision. The quantization effect degrades the accuracy more when the individual class distributions are overlapping in the dataset on which model has converged to optimality. It is harder for the model to differentiate and hence the solution to this can be through hierarchical grouping of similar class distributions. When the object class *dogs* and *cats* are grouped into an object class "small pets", the new object class achieves 90% accuracy, and overall task accuracy rises even at low bit-precision.

| Quantization Approach | Baseline Model (Acc. %) | Post-Quantization at 3-bit | |
|---|---|---|---|
| | | Before Class Grouping | After Class Grouping |
| **Uniform ASYMM** | 90.87 | 85.93 | 88.2 |
| **POWER-OF-2** | 90.87 | 86.9 | 87.7 |

Table 2. Resnet18 CIFAR10 results for Hierarchical Clustering of class distributions with overlapping characteristics. Baseline model is 32-bit.

Table 2 shows the results after performing the hierarchical clustering of the class distributions which are not sufficiently orthogonal for the quantization at lower precision. This example shows that the overall problem statement can be redefined in a quantization-friendly manner for multi-class classification. For example, in autonomous driving tasks, it is not important to differentiate *dogs* vs *cats*, and thus, you don't need to expend extra bits for it.

IV.b. Quantization Effect vs Model Architecture

When the neural networks are similar (e.g. feed-forward convolutional networks), we observed that all the models converge to a similar bit-precision level. Specifically, after uniform quantization approaches give an accuracy drop of less than ~4% up to 6-bit precision before avalanching to very poor results. This is consistent with the Power-of-2 quantization approach as well. The results are summarized in Table 3(a)-3(c).

Table 3(a). Quantization of common DNNs (ImageNet) with ASYMM

| Model Arch./Bit-Width | FP-32 | 8-bits | 7-bits | 6-bits | 5-bits | 4-bits | 3-bits |
|---|---|---|---|---|---|---|---|
| VGG-16 | 71.19 | 71.04 | 70.04 | 60.19 | 0.12 | - | - |
| ResnetV1-50 | 74.08 | 73.94 | 73.33 | 71.17 | 20.57 | - | - |
| ResnetV2-50 | 74.25 | 73.75 | 72.82 | 60.41 | 0.10 | - | - |
| InceptionV2 | 74.43 | 73.92 | 71.72 | 49.97 | 0.13 | - | - |
| InceptionV3 | 76.71 | 76.06 | 72.11 | 52.29 | 0.10 | - | - |
| InceptionV4 | 76.92 | 76.49 | 76.00 | 73.83 | 29.39 | - | - |

Table 3(b). Quantization of Common DNNs (ImageNet) with SYMM

| Model Arch./Bit-Width | FP-32 | 8-bits | 7-bits | 6-bits | 5-bits | 4-bits | 3-bits |
|---|---|---|---|---|---|---|---|
| VGG-16 | 71.19 | 71.15 | 70.87 | 69.47 | 15.63 | - | - |
| ResnetV1-50 | 74.08 | 73.54 | 73.31 | 72.58 | 67.82 | 24.9 | - |
| ResnetV2-50 | 74.25 | 74.22 | 73.33 | 69.63 | 7.88 | - | - |
| InceptionV2 | 74.43 | 74.19 | 73.58 | 68.69 | 30.54 | - | - |
| InceptionV3 | 76.71 | 76.51 | 75.4 | 65.23 | 2.79 | - | - |
| InceptionV4 | 76.59 | 76.46 | 75.95 | 72.42 | 20.02 | - | - |

Table 3(c). Quantization of Common DNNs (ImageNet) with Power-of-2

| Model Arch./Bit-Width | FP-32 | Post-Quantization | Average Bit-Levels |
|---|---|---|---|
| VGG-16 | 71.19 | 64.17 | 5.23 |
| ResnetV1-50 | 74.08 | 69.82 | 5.87 |
| ResnetV2-50 | 74.25 | 70.02 | 6.12 |
| InceptionV2 | 74.43 | - | 5.32 |
| InceptionV3 | 76.71 | - | 6.15 |
| InceptionV4 | 76.59 | 74.34 | 6.15 |

Figure 4 depicts the quantization limit seen across all of the selected ImageNet converged models. From this, we see a "knee of the curve" at 6-bits, suggesting that regardless of model architecture, the quantization limit is closely tied to the dataset. Furthermore, we anticipate that while the different model architectures learn different visual features and semantic associations, at the 6-bit precision, all of the models are limited in their capacity, with respect to object class differentiability.

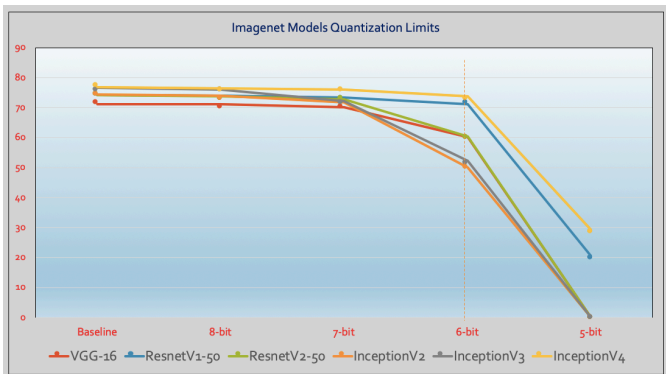

Figure 4. All models show a quantization limit of 6-bit before steep drop in accuracy. The graph shows the ASYMM quantization approach results across models chosen.

IV.c. Quantization Effect on Object Detection Tasks

For evaluation of object detection tasks, we chose the Tiny-YoloV2 model pre-trained on VOC task, and the Mobilenet-SSD pre-trained on MS-COCO task. Table 4 summarizes the results on both the models. We have included a baseline of FP32 accuracy and file sizes. Then, in the middle and right columns, we show results different quantization targets (8-bit and below).

| | Acc | Size MB | Acc | Size MB | −Δ Acc | Δ Size | Acc | Size MB | −Δ Acc | Δ Size |
|---|---|---|---|---|---|---|---|---|---|---|
| Tiny-YOLOv2 VOC 2007 (mAP) | 52.97 | 58.8 (FP-32) | 46.10 | 13.7 (8-bit) | 12.60% | 76.7% | 45.52 | 8.5 (6-bit) | 13.4% | 85.5% |
| MobileNet-SSD Coco dataset (mAP) | 28.56 | 63.28 (FP-32) | 28.22 | 57.12 (Partial 8bit) | 1.19% | 9.73% | 28.09 | 30.48 (Full backbone 8bit) | 1.65% | 51.8% |

Table 4. Quantization results (Uniform ASYMM) on object detection models

Object detection networks generate bounding box coordinates for detected objects. As such, we show results as mAP scores. At low-bit precision, the model must still able to differentiate the object classes as well as identify the location. For Tiny-Yolo, we show a reasonable accuracy loss (12.6-13.4%) for a large gain in compression (76.7-85.5%). We anticipate that the accuracy loss is due to the shallow layers in Tiny-Yolo, which makes it less robust with quantization. Specifically, the MobileNet-SSD model maintains its accuracy within 1-2% loss. We can see that it achieves close to 52% compression benefit without affecting accuracy whilst keeping good precision for the bounding box regressors.

We observed that quantizing the initial layers of the model architecture affects the model performance the most, suggesting (1) the need to preserve the initial learned features, and (2) better results with quantization of only the semantic information layers. Specifically, we show these results for MobileNet-SSD, with partial quantization (partial 8-bits) of the backbone layers and the complete quantization of all backbone.

Results in Table 4 also suggests that the bounding box coordinates in object detection models require a better precision of parameters to regress to a particular real value based on the image coordinates. As such, future quantization schemes may separate the classification and bounding-box components to better compress the model.

IV.d. Compression and Power Benefits

Table 5 shows the file size comparisons using a compression engine (G-zip, 7-zip). In this paper, after we quantize, we store the results back in floating point notation so that we can easily run and compare against baseline FP32. We understand that if we save directly in byte format, we can get 4x savings from the FP32. We show in Table 5 that we can already get 5-6x saving using G-zip compression for the 6-bit quantization (Uniform).

| Model Architecture | Baseline Model Size (MB) | Quantization Approach | Post-Quant Model Size (6-bit) (MB) | | |
|---|---|---|---|---|---|
| | | | G-zip | 7-zip | Δ Size |
| VGG-16 | 513.72 | Uniform | 95.78 | 69.65 | 27.2% |
| | | Power-of-2 | 95.56 | 68.14 | 28.7% |
| ResnetV2-50 | 95.4 | Uniform | 14.05 | 10.12 | 28.0% |
| | | Power-of-2 | 17.84 | 11.18 | 37.3% |
| InceptionV4 | 171.26 | Uniform | 26.6 | 19.4 | 27.1% |
| | | Power-of-2 | 28.6 | 21.6 | 24.5% |

Table 5. Compression numbers using GZip and 7zip

Using logarithmic (power-of-two) quantization, we can get an additional 27%-37% improvement (last column). We find that 7-zip provides better compression for power-of-two, because 7-zip algorithm can better create a dictionary of symbols to compress. In power-of-two, we only have distinct values (1/256…1/4,1/2,…,2,4,8,etc.), and therefore 7-zip's compression approach matches this scheme. It is also important to note that power-of-two scheme has high BE metric, with dense population of values between -1 and 1, with sparser distribution elsewhere to cover a wide range.

|  | Acc | # bit | Size MB | Acc | # Avg.bit | Size MB | −Δ Acc | Δ Size |
|---|---|---|---|---|---|---|---|---|
| Resnet18 CIFAR10 (Acc %) | 90.87 | 32 | 1.1 | 88.90 | 4~5 | 0.15 | 1.9% | 86.6% |
| Tiny YOLOv2 (8 objects) VOC 2007 (mAP) | 52.97 | 32 | 58.8 | 44.72 | 5~6 | 7.7 | 12.2% | 86.9% |

Table 6. Training-Aware Quantization results using Power-of-2

In Table 6, we show a training-aware (power-of-two) quantization results. We see at least an 86% compression using 7-zip, with some reduction in accuracy. With power-of-2 quantization, we can afford reduction in both computational latency and power consumption. For example, we can use bit-shifting instead of multiplications for weights that are powers of two. Processor circuitry using barrel shifters or FPGA logic circuit can easily route values by left or right shifts.

To quantify the saving, we perform simple experiments on both laptop processor (Intel i7, 2.8GHz, 32GB RAM) and a Raspberry PI (ARM Cortex A53, 1.4GHz, 1GB RAM). We ran a tiny CNN model (5-layer LeNet, trained on MNIST, averaged over 10K frame inputs), and we observed 37% and 22% latency improvements from FP32 to integer bit-shifting. Both inference codes were compiled using GCC. Similarly, we observed an energy savings of 37.7% on a Raspberry Pi. We did not measure power on the i7 processor as there are other resident software services in operation. We also know that additional optimization can be implement on both FP32 and integer computation (e.g. using vector instructions and hand-tuned kernels), but the early results shown here are encouraging.

## V. CONCLUSIONS AND FUTURE WORK

We explored various data-free quantization algorithms for a number of deep neural networks (DNN) to better understand quantization limits and their relationships to dataset and model. All of our quantization scheme (asymmetric, symmetric, power-of-two) achieves near original model accuracy for every model we tested. This provides flexible hardware configuration that uses hardware-bound instructions to reduce power and size. The variation of quantization may also imply robust sparsity encoding of the neural networks.

Using ImageNet as a reference dataset, we find that the selected state-of-art DNN all converge similarly around an average 6-bit precision. We anticipate this is due to a number of factors, including: (1) the DNNs have similar feed-forward convolutional layer that essentially learn similar features, (2) the DNNs have sufficient capacity and depth such that they manage to learn the dataset distribution. We find that the best quantization results come from the ability to both provide the range and distribution of bits needed by the model to learn dataset distribution. We show that we can improve bit precision beyond quantization limits with simple object class clustering.

Going forward, there is much more we can do as future work to extend the quantization schemes. We are exploring ways to automate the post-training quantization schemes to not only compress the parameters, but also to visualize the results so that the AI developer can best chart out the next steps. We are exploring means to automate the selection of the quantization schemes, in addition to chaining the processes up (e.g. symmetric followed by power-of-two) because successive quantization steps may tease out different qualities of the parameter variations with respect to range and density. We are also exploring regression quantization that separates the classification and bounding-box components.

## VI. ACKNOWLEDGEMENTS

We acknowledge the support/effort by the Latent AI and SRI teams: Vasil Daskalopoulos, Mark Griffin, Indu Kandaswamy, Joe Zhang, Saurabh Farkya and Aswin Raghavan.